\documentclass[letterpaper, 10 pt, conference]{ieeeconf}  % Comment this line out if you need a4paper

\IEEEoverridecommandlockouts                              % This command is only needed if 
\usepackage{amsmath,amssymb}
\usepackage{graphicx,xcolor}
\usepackage{algorithm}
\usepackage{algpseudocode}
\usepackage{verbatim}

\title{\LARGE \bf
Suture Thread Modeling Using Control Barrier Functions for Autonomous Surgery 
}

\author{Kimia Forghani\textsuperscript{1}, Suraj Raval\textsuperscript{1}, Lamar Mair\textsuperscript{2}, Axel Krieger\textsuperscript{3}, and Yancy Diaz-Mercado\textsuperscript{1}% <-this % stops a space
\thanks{\textsuperscript{1} Authors are with the Department of Mechanical Engineering,
        University of Maryland, College Park, MD 20742
        {\tt\small (kimiaf, sraval, yancy) @umd.edu}}%
\thanks{\textsuperscript{2} Author is with the Division of Magnetic Manipulation and Particle Research, Weinberg
Medical Physics, Inc., North Bethesda, MD 20852
        {\tt\small lamar.mair@gmail.com}}%
\thanks{\textsuperscript{3} Author is with the Department of Mechanical Engineering, Johns Hopkins University,
Baltimore, MD 21218
        {\tt\small axel@jhu.edu}}%
\thanks{“© 2025 IEEE.  Personal use of this material is permitted.  Permission from IEEE must be obtained for all other uses, in any current or future media, including reprinting/republishing this material for advertising or promotional purposes, creating new collective works, for resale or redistribution to servers or lists, or reuse of any copyrighted component of this work in other works.”  } 
}

\begin{document}

\maketitle
\thispagestyle{empty}
\pagestyle{empty}

%%%%%%%%%%%%%%%%%%%%%%%%%%%%%%%%%%%%%%%%%%%%%%%%%%%%%%%%%%%%%%%%%%%%%%%%%%%%%%%%
\begin{abstract}

 Automating surgical systems enhances precision and safety while reducing human involvement in high-risk environments. A major challenge in automating surgical procedures like suturing is accurately modeling the suture thread, a highly flexible and compliant component. Existing models either lack the accuracy needed for safety-critical procedures or are too computationally intensive for real-time execution. In this work, we introduce a novel approach for modeling suture thread dynamics using control barrier functions (CBFs), achieving both realism and computational efficiency. Thread-like behavior, collision avoidance, stiffness, and damping are all modeled within a unified CBF and control Lyapunov function (CLFs) framework. Our approach eliminates the need to calculate complex forces or solve differential equations, significantly reducing computational overhead while maintaining a realistic model suitable for both automation and virtual reality surgical training systems. The framework also allows visual cues to be provided based on the thread’s interaction with the environment, enhancing user experience when performing suture or ligation tasks. The proposed model is tested on the MagnetoSuture system, a minimally invasive robotic surgical platform that uses magnetic fields to manipulate suture needles, offering a less invasive solution for surgical procedures.  

\end{abstract}

%%%%%%%%%%%%%%%%%%%%%%%%%%%%%%%%%%%%%%%%%%%%%%%%%%%%%%%%%%%%%%%%%%%%%%%%%%%%%%%%
\section{INTRODUCTION}
Recent advancements in surgical robotics have underscored the potential of autonomous systems to revolutionize healthcare by enhancing precision, reducing complications, and ensuring consistent surgical outcomes \cite{reddy2023advancements}. The COVID-19 pandemic further emphasized the need to minimize human interaction in high-risk environments. Autonomous systems address challenges like infection risks, communication delays in telesurgery, and are ideal for tasks like suturing due to their repetitive nature. The success of robotic platforms like the da Vinci system reflects the medical community's readiness to embrace these technologies
% , underscoring their crucial role in the future of surgery 
\cite{attanasio2021autonomy}. 

To achieve safe, fully autonomous surgeries, accurate modeling of all components is instrumental. Understanding suture thread behavior is a particularly important yet understudied component in surgical robotics \cite{ostrander2024current}. This is because cable-like objects, such as threads, wires, and hair, are challenging to model due to the inherent trade-off between accuracy and real-time performance \cite{lv2020review}. However, proper suture thread modeling is key for simulating and performing tasks like suturing, knot-tying, and ligation. Beyond surgery, this model has applications in virtual reality training, textile simulations, hair or cable animation, and soft robotics, where realistic thread behavior is critical.

Recent advances in modeling suture threads are mainly based on continuum mechanics approaches, finite element methods \cite{jourdes2022visual}, position-based dynamics (PBD) approaches  \cite{yu2020real,moore2023interactive}, or a combination of them \cite{qi2017virtual,xu2018real}. Continuum mechanics and finite element methods provide high accuracy, but are computationally intensive due to the need to solve complex equations governing deformation on top of applying constraints like inextensibility. Additionally, collision detection and contact forces further increase computational demand, especially for real-time simulations. PBD relies on solving constraints iteratively in a Gauss-Seidel fashion to estimate the thread's position in each iteration. PBD does not solve all constraints simultaneously but adjusts positions gradually through successive approximations. This allows for quick convergence, especially in systems with many constraints, making PBD computationally efficient. However, the order in which constraints are applied can significantly impact the results, potentially leading to oscillations especially in over-constrained scenarios. Although PBD is computationally efficient and thus favored in applications like computer graphics and video games, it sacrifices physical accuracy for speed.

Control barrier functions (CBFs) are a mathematical framework used in control theory to enforce constraints while ensuring system stability and safety \cite{ames2019control}. They can be formulated as a quadratic program (QP) for fast online computation of safe control inputs. Unlike PBD, CBFs ensure that the constraints are met by influencing the control actions of the system, not by directly manipulating the positions of the entities. Due to their safety assurances, barrier certificates have been applied to various problems, including collision avoidance for autonomous agents \cite{jankovic2021collision,liu2023clf}, adaptive cruise control \cite{chinelato2023design}, and lane keeping \cite{bruggemann2022simultaneous}.

In this paper, we propose using CBFs to model suture threads both accurately and efficiently. We model the thread as a line graph consisting of $n$ single integrators, whose velocities can be directly controlled and adjusted to meet all safety constraints simultaneously. This is achieved through solutions to a QP, which benefits from the inherent sparsity of the constraints in a line graph structure. By leveraging this property, the computational cost is significantly reduced. This approach allows us to model connectivity, stiffness, damping, and obstacle interactions, without the need to solve Newtonian equations or sacrifice accuracy. The model is validated experimentally using the MagnetoSuture system, a robotic platform that magnetically controls needle position, enabling tetherless minimally invasive surgeries.

The outline for the rest of the paper is as follows: Section 2 provides background on CBFs, and Section 3 explains the suture thread model utilizing CBFs and CLFs. In Section 4, the accuracy of the suture thread model is validated through comparisons between model-based simulations and experimental robot data of suture thread movement. Additionally a surgical task simulation with visual feedback is provided.

\section{Background}
In this section, we cover the basics of system safety and control, beginning with a control model and introducing CBFs for enforcing safety. We then explain how CBFs can be integrated into a QP to ensure optimized safe operation.
\subsection{System and Safety}

Consider an affine control system:
\begin{equation}
\dot{x} = f(x) + g(x)u,
\end{equation}
with \(x \in \mathbb{R}^d\) and \(u \in \mathbb{R}^m\), where \(f\) and \(g\) are locally Lipschitz. The system is subject to input constraints \(u(t) \in\mathcal{U}\subseteq\mathbb{R}^m\), the space of admissible control signals.

A set \(C \subset \mathbb{R}^d\), defined by:
\begin{equation}
\begin{array}{l}
C = \{x \in \mathbb{R}^d : h(x) \geq 0\}, \\
\partial C = \{x \in \mathbb{R}^d : h(x) = 0\}, \\
\text{Int}(C) = \{x \in \mathbb{R}^d : h(x) > 0\}.
\end{array}
\end{equation}
is forward invariant if, under a feedback controller \(\pi(x, t)\), the solution \(x(t)\) of the closed-loop system:
\begin{equation}
\dot{x} = f(x) + g(x)\pi(x, t),
\end{equation}
remains in \(C\) for all time. The system is safe with respect to \(C\) if \(C\) is forward invariant \cite{ames2019control}.

\subsection{CBFs and CLFs}

A control barrier function (CBF) ensures system safety by keeping the state within a safe set defined by a function \(h(x)\). The function \(h\) is a CBF if there exists a class \(\mathcal{K}_\infty\) \cite{sastry2013nonlinear} function \(\alpha\) such that:
\begin{equation}
    \sup_{u \in \mathcal{U}} \dot{h}(x, u) \geq -\alpha(h(x)),
\end{equation}
for all \(x\) in the domain. This condition imposes a lower bound on the evolution of $h$ through the control policy so the system remains in the safe set and maintains safety over time.

In addition to ensuring safety, we are also interested in stabilizing the system to a desired state. This can be achieved using a control Lyapunov function (CLF), which drives the system to stability. A function \(V(x)\) is a CLF if it is positive definite and there exists a class \(\mathcal{K}_\infty\) function \(\gamma\) such that
\begin{equation}
\label{eq:CLF}
    \inf_{u \in \mathcal{U}} \dot{V}(x, u) \leq -\gamma(V(x)),
\end{equation}
for all \(x\) in the domain. This condition imposes an upper bound on the evolution of $h$ through the control policy, driving the system towards the desired equilibrium state over time.

\subsection{CBF-Based Control}

Given a feedback controller \( u = k(x) \) for a control system, we may encounter situations where \( k(x) \) does not ensure safety according to the CBF criteria. Thus, we leverage the fact that the safety condition, expressed as:
\begin{equation}
\label{eq:safety_condition}
\dot{h}(x,u) = L_f h(x) + L_g h(x) u \geq -\alpha(h(x)),
\end{equation}
is affine in \( u \), where \( L_f h(x) \) and \( L_g h(x) \) are the Lie derivatives \cite{sastry2013nonlinear} with respect to \( f(x) \) and \( g(x) \), respectively. This allows us to formulate a quadratic program (QP) that finds the control input \( u(x) \) with the smallest deviation from the desired control \( v(x) \) \cite{ames2019control}: 
\begin{equation}
\begin{gathered}
    u(x) = \arg\min_{u \in \mathbb{R}^m} \frac{1}{2} \|u - v(x)\|^2\\
    \begin{aligned}
        \text{s.t.} \quad &\dot{h}(x_k, u_k) \geq -\alpha_k h(x_k)  \\
        & \dot{V}(x, u) \leq -\gamma(V(x)).
    \end{aligned}
\end{gathered}
\end{equation}
% 
% subject to the safety constraint in \eqref{eq:safety_condition}.
This QP-based approach ensures that the system remains within the safe set by minimally perturbing the control input. The same QP is also subject to the constraint in \eqref{eq:CLF}, which drives the system to stability. This type of controller is referred to as CLF-CBF-QP.

\begin{figure}[b]
      \centering
     
      \includegraphics[width=3.2in]{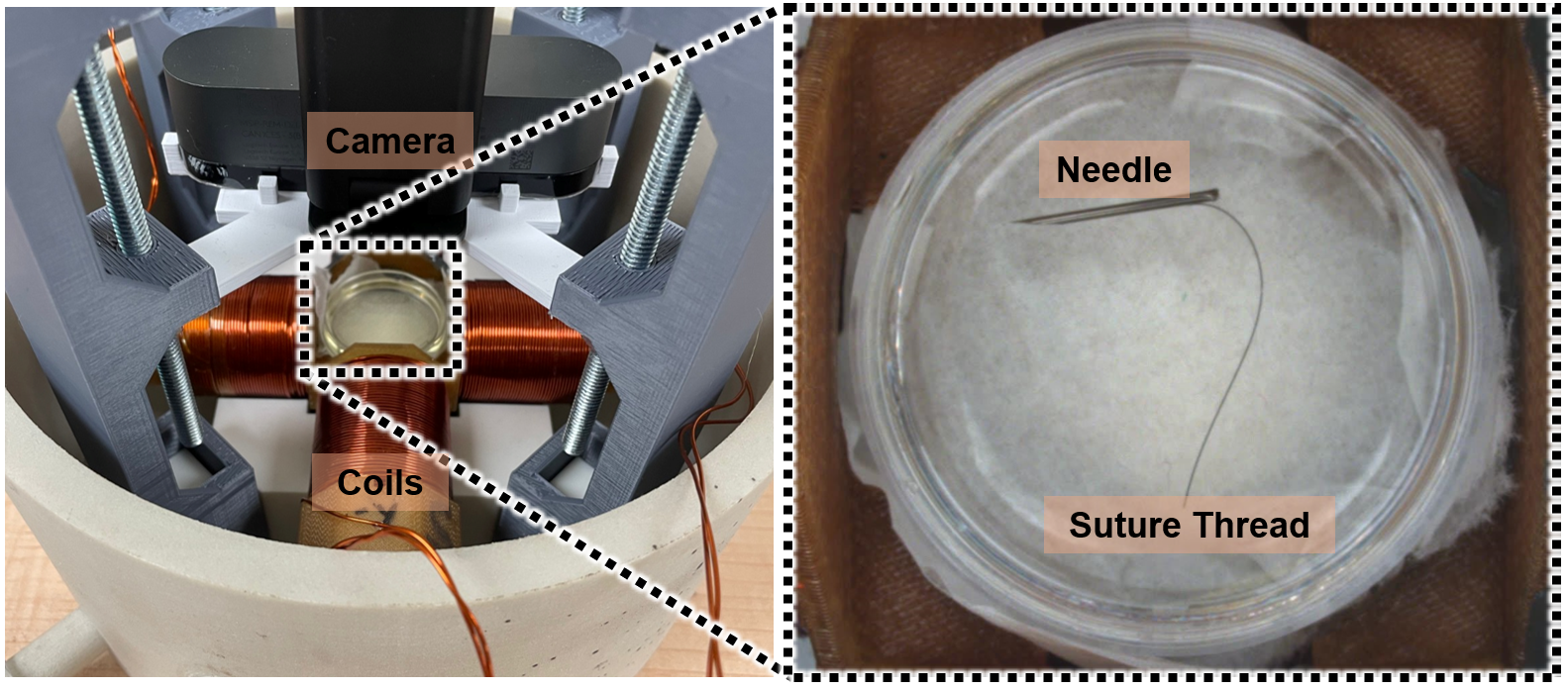}
    
      \caption{The MagnetoSuture system (left), and closed up look at the workspace that includes the suture needle and thread (right). MagnetoSuture is a magnetic manipulator robot which leverages vision-based feedback to control magnetic tools (e.g., needle) via a coil-generated magnetic field. The workspace used in this setup is a petri dish, 35mm in diameter.}
      \label{fig:magneto}
   \end{figure}

\subsection{MagnetoSuture}
The experiments included in this paper are performed by the MagnetoSuture \cite{mair2020magnetosuture} system as shown in Fig.~\ref{fig:magneto}. This surgical robotic system enables tetherless manipulation of suture needles using magnetic fields, eliminating the need for large robotic tools during minimally invasive surgery. This approach reduces invasiveness, as the only components introduced into the body are the needle and thread, lowering the risk of tissue damage, scarring, and infections. The system uses an electromagnetic coil array to guide NdFeB (neodymium iron boron) suture needles for tasks like tissue penetration, ligation, and complex suture patterns demonstrated in successful experiments with \textit{ex vivo} tissues.

\section{Modeling Suture Thread with Control Barrier Functions}

\begin{figure}[b]
      \centering
           \includegraphics[width=3.2in]{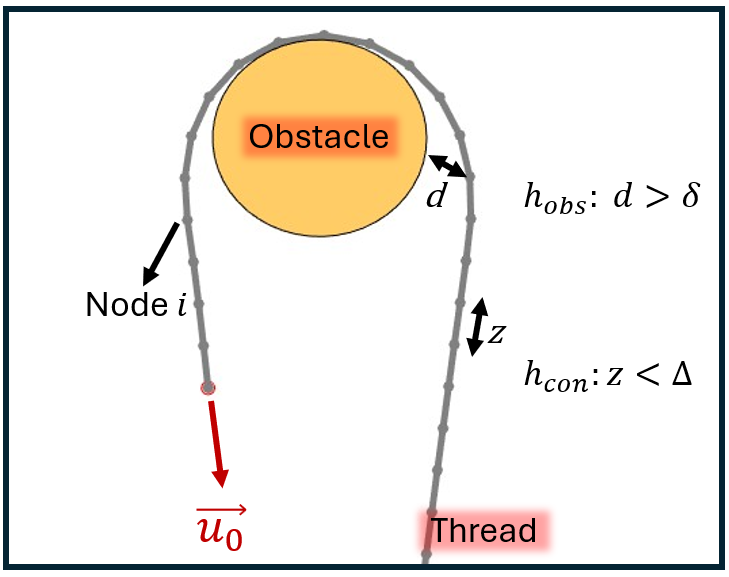}\vspace{-1em}
          \caption{Conceptual representation of the suture model, moving with needle velocity of \(u_0\), and interacting with a circular obstacle. In the figure, $z$ corresponds to the distance between nodes, and $d$ is the shortest distance from a node to the obstacle boundary. The constraints encoded in $h_{obs}$ and $h_{con}$ enforce that the thread does not enter the obstacle and that it does not extends past a maximum length, respectively.}
      \label{fig:model}
   \end{figure}

In this section, we model the suture thread as a system of \(n\) nodes and a lead node (needle), each represented as a single integrator. The state dynamics of each node are:
% given by:
% 
\begin{equation}\label{eq:dyn}
\begin{aligned}
  \dot{x}_i &= u_i, \quad x_i \in D\subset\mathbb{R}^d, \quad i = 1, \dots, n, \\
  \dot{x}_0 &= u_0, \quad x_0 \in D\subset\mathbb{R}^d
\end{aligned}
\end{equation}
Here, \( \dot{x}_i \) represents the velocity of the \(i\)-th node, and \( u_i \) is the control input. The thread's lead node corresponds to the surgical needle, with \( \dot{x}_0 \) and \( u_0 \) representing its velocity and control input, respectively. The nodes' velocities (\( u_0, \dots, u_n \)) are determined using a QP, which minimizes deviations from a desired velocity while ensuring the system remains in a safe state. To guarantee forward invariance, the initial states of all nodes, (\( x_0, \dots, x_n \)), are chosen within the safety set. The desired control input for the lead node can either follow a predefined path or be adjusted in real-time by the user via a joystick. The unsafe states are described by the CBFs discussed in this section. Fig.~\ref{fig:model} shows a conceptual figure of the thread model and its constraints.

\subsection{CBF for Connectivity}
This CBF ensures that the natural connectivity of the thread is maintained by enforcing that the distance between any two consecutive nodes does not exceed a maximum allowable distance. We note that when the thread loops, it is possible for two nodes to become coincident. For this reason, we do not impose a minimum distance constraint. 

For convenience, we define the ensemble vector of position states as \( x = [x_1^T, \dots, x_n^T]^T \in {R}^{nd} \), and the ensemble vector of control velocities as \( u = [u_1^T, \dots, u_n^T]^T \in {R}^{nd} \). 
For $i = 2, \dots, n$, we define a connectivity function as \cite{li2018formally}:
\begin{equation}
    h_{\text{con},i}(x) = \tfrac{1}{2} (\Delta^2 - \|x_i - x_{i-1}\|^2),
\end{equation}
where \( \Delta > 0 \) is the maximum node separation distance. 

 The connectivity CBF is applied only to the states of the thread nodes and does not influence the control input of the lead node. However, to ensures that the other nodes follow the needle while maintaining thread connectivity, we introduce an additional connectivity function related to the needle's position \( x_0 \):
\begin{equation}
\label{hcon1}
h_{\text{con,1}}(x, x_0) = \tfrac{1}{2} (\Delta^2 - \|x_1 - x_0(t)\|^2),
\end{equation}
The safe state \( C_{\text{con,i}} \)
% , corresponding to the connectivity function 
is defined as:
\begin{equation}
C_{\text{con,i}} = \{x \in D \mid h_{\text{con,i}}(x) \geq 0\}.
\end{equation}
The barrier function in \eqref{hcon1} ensures that the lead node, or needle, maintains connectivity with the rest of the thread. However, since the model is implemented in discrete time, depending on the user input, at each time step, the needle can be positioned at a distance from the first node such that it exits the safe state. In this case, the invariance of set \( C \) isn't guaranteed because we are not starting from a safe state. However, even if the needle is positioned in an unsafe state (i.e., \( h_{\text{con,1}} < 0 \)), after a few iterations, the system controls the rest of the nodes to converge into a safe state. The smaller the deviation of the needle's position from the safe state is, the faster the convergence will be. To account for this error, slack variables are introduced as discussed in Section \ref{subs:cont}.

Additionally, we enhance connectivity, by linking each node $i = 3, \dots, n$, not only to node \(i-1\) but also to node \(i-2\). This approach increasing the algebraic connectivity (the second smallest eigenvalue of the graph Laplacian matrix) which has been shown to enhance system robustness \cite{jamakovic2008robustness}.
% By establishing additional connections, we increase \(\lambda_1\), thereby improving the model’s stability and making it more resistant to perturbations. 
% This structural enhancement is critical for achieving more realistic and reliable simulations.
For $i = 3, \dots, n$, we define this connectivity function as:
\begin{equation}
\begin{aligned}
    h_{\text{con enhanced},i-1}(x) &= \tfrac{1}{2} (\Delta^2 - \|x_i - x_{i-2}\|^2), \\
    h_{\text{con enhanced},1}(x) &= \tfrac{1}{2} (\Delta^2 - \|x_2 - x_0(t)\|^2).
\end{aligned}
\end{equation}
The safe state \( C_{\text{con enhanced,i}} \), corresponding to this connectivity function is defined as:
\begin{equation}
C_{\text{con enhanced,i}} = \{x \in D \mid h_{\text{con enhanced,i}}(x) \geq 0\}.
\end{equation}

\subsection{CBF for Obstacle Avoidance}
This section presents a CBF that maintains a minimum distance between each node and the nearest obstacle, preventing penetration through the obstacle.
%This section discusses a CBF that ensures that the distance between each node and the nearest obstacle remains greater than a minimum allowable distance, preventing the thread from penetrating the obstacle.% while allowing contact with its surface. 
% 
%When enforcing inequality constraints on the distance function between the robot and obstacles, these components are usually approximated as ellipsoids \cite{ferraguti2020control} or hyper-spheres \cite{zeng2021safety}. This is mainly because sharp corners in obstacles cause non-differentiability due to abrupt changes in the closest point to the robot. As the robot moves near a sharp corner, the closest point can switch suddenly from one edge to another, leading to an abrupt change in the direction of the connecting vector, making the distance function non-differentiable at these points. The advantage of approximating these components as curved shapes is that the distance functions for these shapes are differentiable and can serve as control barrier functions to formulate a safety-critical optimal control problem. However, these approximations often overestimate the size of both the robot and the obstacles.%
When enforcing distance-based constraints, obstacles are often approximated as ellipsoids \cite{ferraguti2020control} or hyper-spheres \cite{zeng2021safety} to avoid non-differentiability caused by sharp corners. As the robot approaches a corner, the closest point can shift abruptly between edges, causing discontinuities in the distance function. Approximating these components as curved shapes ensure differentiability, enabling use of control barrier functions. However, they tend to overestimate obstacle and robot sizes.

\begin{figure}[t]
      \centering
      \framebox{
      \includegraphics[width=3in]{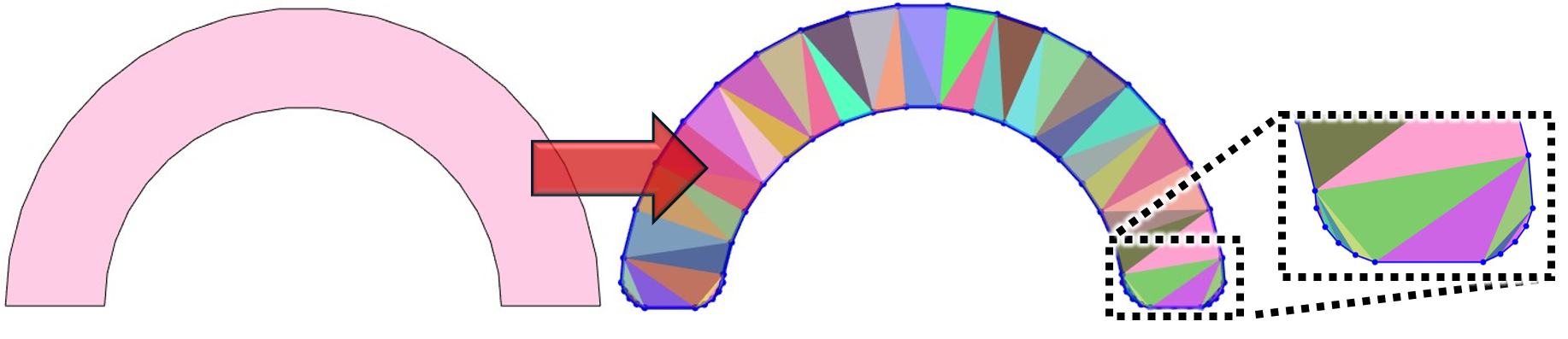}
      }
      \caption{Smoothing and triangulation of a non-convex shape (left) using the Delaunay method. The polygon vertices and edges are depicted in blue, while the triangulation is highlight by  arbitrary colors (right).}
      \label{fig:tri}
   \end{figure}

To mitigate this, we propose a method which is applicable to both convex and non-convex obstacles. In this approach we smooth the corners by estimating curves that approximate only the sharp regions, reducing the abrupt direction changes while keeping the integrity of the obstacle shape. Although the distance function remains non-differentiable, the transition between edges becomes much smoother, and the changes in direction are less drastic. Then the obstacles are divided into triangles using the Delaunay triangulation method \cite{fortune2017voronoi}. Fig.~\ref{fig:tri} illustrates an examples of a non-convex shape that has been smoothed and triangulated.% using this approach. 

Furthermore, for each node, the closest point on each triangle is determined by projecting the node’s position onto the triangle’s edges and clamping the projection to ensure the closest point remains on the edge. The overall closest point on the obstacle is then selected by finding the minimum distance across all triangles, and this distance is used to compute the CBF below. This method efficiently handles large sets of points and triangles by minimizing loop usage and relying on vectorized operations.

For nodes $i = 1, \dots, n$ and obstacles $O = 1, \dots, M$, we define an obstacle avoidance function as:
\begin{equation}
h_{\text{obs,}i, O}(x, p) = \tfrac{1}{2} (\|x_i - p_O\|^2 - \rho^2),
\end{equation}
where \(p_O\) is the position of the closest point on obstacle \(O\) to node \(i\), and \(\rho\) is the minimum allowable distance between the two points. 
Similarly, we define an obstacle avoidance function for the lead node as:
\begin{equation}
h_{\text{obs,needle,} O}(x, p) = \tfrac{1}{2} (\|x_0 - p_O\|^2 - \rho^2),
\end{equation}
As opposed to the connectivity CBF, the obstacle avoidance CBF affects the needle control input to stop it from penetrating obstacles, despite the user input or a defined path.

The safe state \( C_{\text{obs}} \), corresponding to this collision avoidance function is defined as:
\begin{equation}
C_{\text{obs},i} = \{x \in D \mid h_{\text{obs,i}}(x) \geq 0\}.
\end{equation}

\subsection{CLF for Stiffness}
For suture threads, which typically exhibit memory effects, the thread naturally tends to return to its original shape when no additional forces are provided. This behavior is modeled using a CLF rather than a CBF, as the thread has an equilibrium state it seeks to maintain, rather than a safe set to stay within. The memory is modeled as the distance between node \(i\) and node \(i-2\) for \(i = 3, \dots, n\) in the natural state. For example, if the memory of the suture thread is for it to be a straight line, then the natural state distance between nodes would be \(2\times \Delta\), where \(\Delta\) is the maximum separation distance between two adjacent nodes. 

\begin{figure}[t]
      \centering
      \framebox{
      \includegraphics[width=3in]{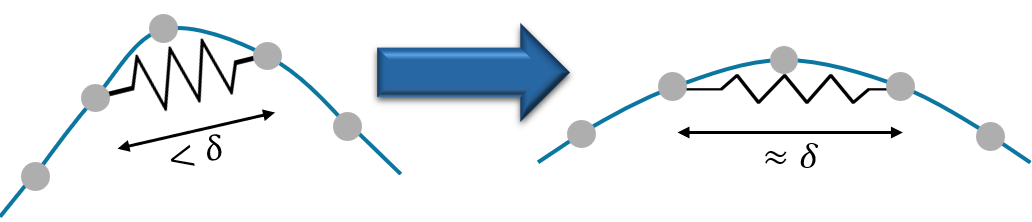}
      }
      \caption{The CLF in \eqref{stiffclf} acts analogous to a stiff mechanical spring between nodes \(i\) and \(i+2\). Nodes are represented by gray circles.}
      \label{fig:delta}
   \end{figure}

For the general case, the CLF is defined as:
\begin{equation}
\label{stiffclf}
\begin{array}{cc}
     & V_{i-1}(x) = \tfrac{1}{2} \left( \|x_i - x_{i-2}\|^2 - ({\delta_i})^2 \right)^2, \\
     & V_{1}(x) = \tfrac{1}{2} \left( \|x_2 - r(t)\|^2 - ({\delta_2})^2 \right)^2,
\end{array}
\end{equation}
where \(\delta_i\) is distance of the respective 2 nodes in the natural state. Using this CLF in the controller ensures that the thread naturally returns to its original shape over time. This concept is equivalent to having a stiff spring attached between node \(i\) and \(i+2\), whose equilibrium state is achieved when the two nodes are at distance \(\delta\). Fig.~\ref{fig:delta} illustrates when the thread's curvature exceeds its natural state, the `spring' compresses, pushing the nodes apart to restore its natural state. Conversely, when the nodes are too far apart, the spring stretches, pulling them back toward the natural equilibrium, maintaining the thread's overall shape.

\subsection{Damping}
Up until this point, it is assumed that the thread moves on a high-friction surface, which implies that, in the absence of control input, the thread remains stationary. However, in our experimental setup, the thread floats on a viscous fluid as part of the MagnetoSuture system. In this case, 
% the previous assumption no longer holds, and 
a new damping assumption is required. 

 Instead of assuming that the desired velocity of each node, which we optimize to minimize deviation from, is zero, we assume it is a fraction \(\kappa\) of the node's current velocity. 
 % The remaining 5\% accounts for the minimal damping of the viscous medium that the thread moves in. 
The damping percentage was determined experimentally to be \(\kappa=0.95\) by tuning the simulation to match the experimental results of the thread moving in glycerin. 
%Using this setup highlights the flexibility of our approach in modeling the dynamics of the suture thread in various environments.

\subsection{Controller}
\label{subs:cont}
The overall thread safety set is the intersection of all $L=3n-2 + M(n+1)$ safety sets which include: $n$ connectivity constraints,  $n-1$ enhanced connectivity constraints, $M(n+1)$ obstacle constraints, and $n-1$ stiffness constraints:
\begin{equation}
%C_{\text{thread}} = \bigcap_{\ell=1}^n C_{\ell,i}.
C_{\text{thread}} =\bigcap\nolimits_{\ell=1}^{3n-1 + M(n+1)} C_{\ell}
\end{equation}
The control input is determined by solving the following QP:
\begin{gather}
\label{eq:control}
\min_{\substack{u \in \mathbb{R}^{nd}, u_0\in\mathbb{R}^d\\s\in\mathbb{R}^{3n-2}}} \tfrac{1}{2} \|u - v(x)\|^2 +\tfrac{1}{2} \|u_0 - v_0(t)\|^2 + \phi(s) \\
\text{s.t.}\qquad
A_{\text{obs}}\,[u_0^T,u^T]^T\geq b_{\text{obs}}\nonumber\\
A_{\text{con}}\,u\geq b_{\text{con}}-[s_1,\ldots,s_{n}]^T \nonumber\\
A_{\text{con,enh}}\,u\geq b_{\text{con,enh}}-[s_{n+1},\ldots,s_{2n-1}]^T \nonumber\\
A_{\text{stiff}}\,u\leq b_{\text{stiff}}+[s_{2n},\ldots,s_{3n-2}]^T\nonumber\\
s\geq0\nonumber
\end{gather}
Here, the desired value for node and needle velocities, \(v(x)\) and \(v_0(x)\) are adjusted based on the environment damping properties and the user input. The slack variables \( s^T = [s_1, \dots, s_{3n-2}] \) are introduced to relax connectivity and stiffness constraints, ensuring the existence of a feasible solution even when constraints must be violated due to the needle’s input. Note that even if this is the case, the properties of the CLF-CBF-QP eventually drive the system back to the safe region. Additionally, the slack penalty function $\phi(s) = \frac{1}{2} s^T W s$,
where \( W \) is a diagonal matrix of positive weights, enables constraint prioritization. Connectivity constraint is heavily enforced with \( W_{\text{con}} \gg W_{\text{stiff}} \). Stiffness constraint is assigned lower weights, empirically tuned based on suture material. Obstacle avoidance is treated as the highest priority, with no slack allowed. The cost function is normalized such that \( W = 1 \) for terms that minimize velocity deviations.

\begin{figure}[t]
      \centering
      \includegraphics[width=3.2in]{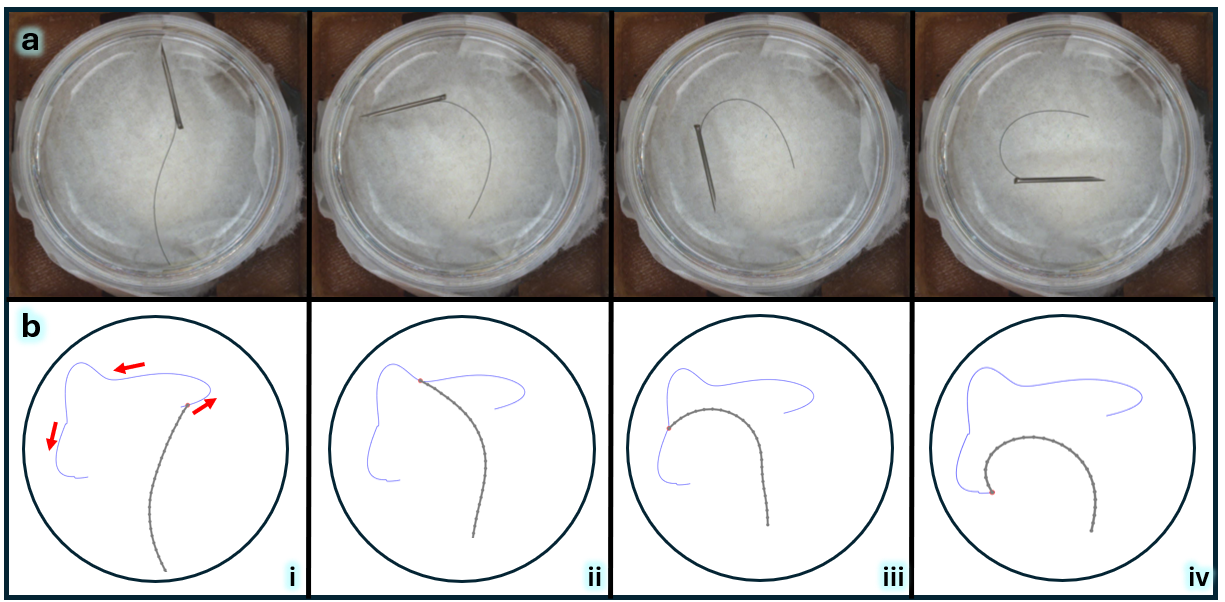}
           \caption{Visual comparison of (a) the experimental behavior of a 19 mm long Polyamide suture thread in the MagnetoSuture system and (b) the behavior of the proposed model for four instances in time (i-iv). For simplicity, the needle is not depicted; instead, its endpoint is represented by a red circle around the lead node.}
      \label{fig:exp}
   \end{figure}   

Due to our simplified dynamics, the entries for the constraint matrices are given by partial derivatives and evaluation of the constraints functions. For example, for the $i$\textsuperscript{th} connectivity constraint function, we get
\begin{align}
    [A_{con}]_{i,j} = \frac{\partial h_{i,con}}{\partial x_j}, &&& [b_{con}]_i = -\alpha_k h_{con,i}
\end{align}
As these functions only rely on local state information (e.g., $x_i$ for $i=1,...,n$, and $x_{i-1}$ for $i=2,...,n$), the majority of the entries in $A_{con}$ will be zero. This is true for the other constraints as well. Thus, for \(U^T=[u^T,u_0^T,s^T]\), it is possible to represent all the constraints in \eqref{eq:control} as a single highly sparse inequality
% 
% If we rewrite the conditions in \eqref{eq:control}, it is equivalent to having:
\begin{equation}
\label{eq:axb}
    AU + b \geq 0,
\end{equation}
which significantly reduces the computational complexity of this optimization task.

\section{Results}
In this section, experimental maneuvering tasks were conducted on the MagnetoSuture system to compare with the proposed simulation model, and both robotic and simulated tasks were video-recorded to assess suture thread behavior.

      \begin{figure}[t]
      \centering
      \includegraphics[width=3.4in]{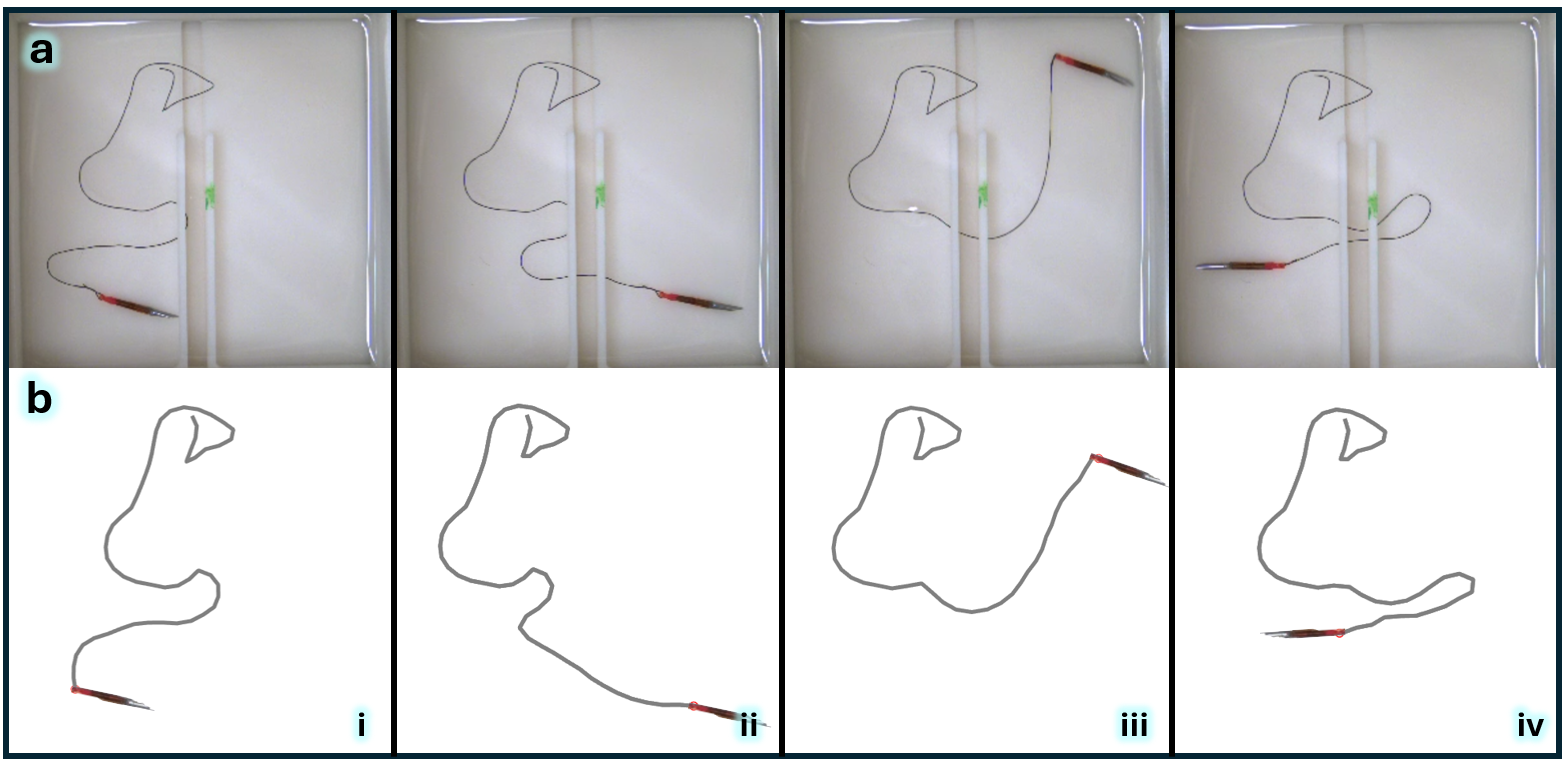}
            \caption{Visual comparison of (a) the experimental behavior of a 150 mm long silk suture thread in the MagnetoSuture system and (b) the behavior of the proposed model. The needle shown in (b) is a visual copy of the needle from (a), to improve clarity, and is not part of the simulation output.}
      \label{fig:silk}
   \end{figure}
   
   \begin{figure}[b]
      \centering
      \includegraphics[width=3.4in]{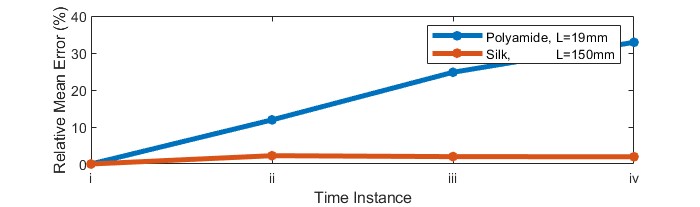}\vspace{-1em}
            \caption{Mean error percentage relative to thread length for two experiments. Number of simulated nodes is  $n=25$ (Polyamide) and $n=81$ (Silk).}
      \label{fig:meanerror}
   \end{figure}

   \begin{figure}[t]
      \centering
      \includegraphics[width=3.4in]{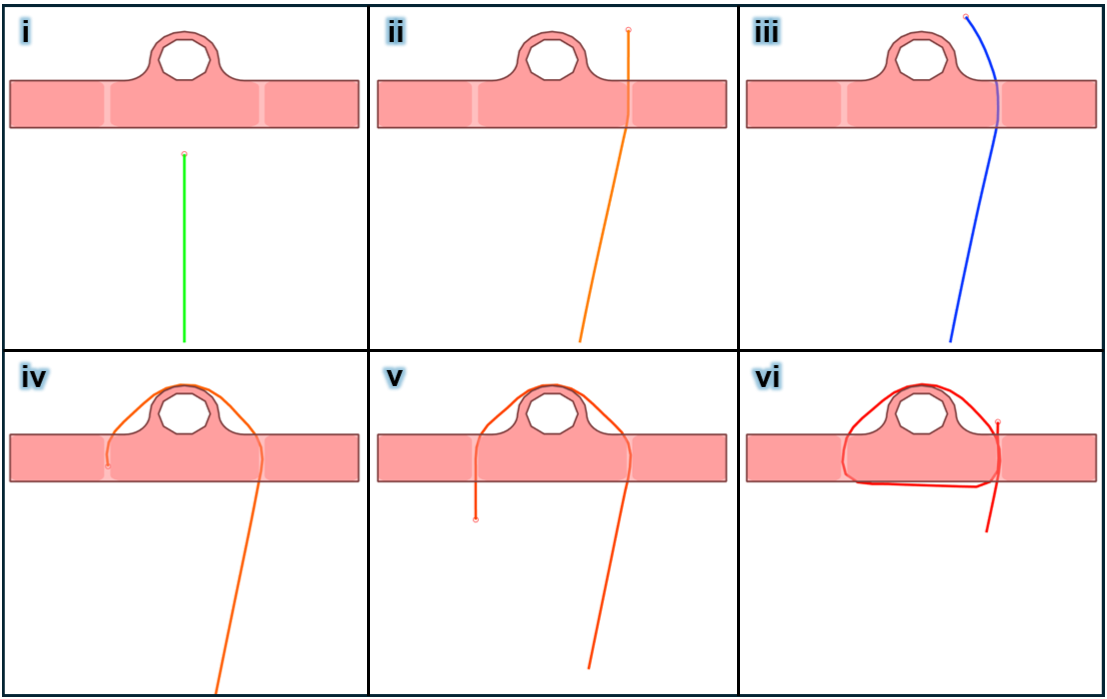}\vspace{-1em}
            \caption{Six time instances from the real-time simulation of an inguinal hernia repair using the proposed suture thread model. The pink geometry with a ring is the inguinal hernia in tissue, and the colored line is the suture thread with 40 nodes. i) The thread moves freely towards the obstacle (T=0s). ii) The thread turns orange, indicating friction in the penetration path, slowing the needle (T=12s). iii) The lead node moves left, not pulling the part of the thread under friction, turning it blue (T=17s). iv-vi) The thread wraps around the obstacle, changing color and slowing further as friction increases (T=27s, 35s, 55s).}
      \label{fig:hernia}
   \end{figure}

   \begin{figure}[t]
      \centering
      \includegraphics[width=3in]{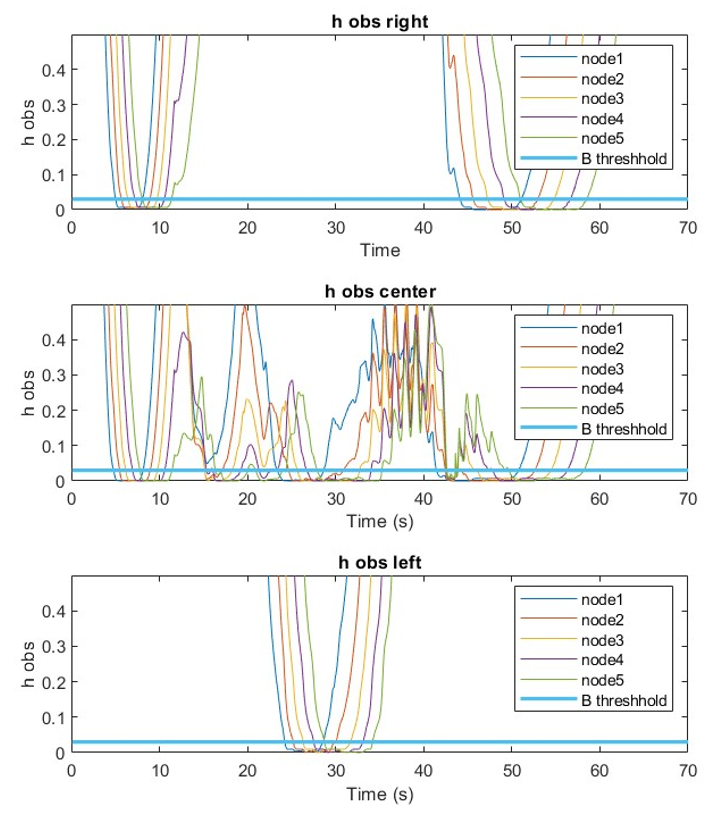}\vspace{-1em}
            \caption{The obstacle constraint function \(h_{obs}\) corresponding to the first five nodes of the thread for the three obstacles used in the hernia simulation.}
      \label{fig:hobs}
   \end{figure}

\subsection{Thread Behavior Test}
The proposed model is validated against experiments using two suture threads: a USP 7-0 polyamide monofilament and a USP 6-0 silk braided suture, both maneuvered in a glycerin medium. The silk thread was tested in a larger workspace to assess adaptability. These threads were selected to evaluate the model’s ability to capture varying stiffness, as polyamide is inherently stiffer than silk. Fig. \ref{fig:exp} and Fig. \ref{fig:silk} compare simulated and experimental thread behavior along arbitrary paths. The simulation runs at 66 Hz.

Fig.~\ref{fig:meanerror}  illustrates the mean error relative to the thread length for various lengths and materials. The silk suture thread exhibits minimal error, primarily due to two factors. First, silk has lower inherent stiffness compared to polyamide, making it easier to model accurately. Second, the method of attachment to the needle affects performance. The polyamide thread is glued to the needle, creating increased rigidity near the needle, causing the thread to assume a specific angle that the model does not account for (this is not the conventional method of attachment). In contrast, the silk thread is passed through a hole in the needle, allowing it to move freely. Future experiments will involve silk sutures performing tasks such as suturing for further model validation.

\subsection{Hernia Repair Simulation with Visual Feedback}
We use the proposed model to simulate an inguinal hernia repair procedure. The hernia is modeled as a ring with an outer diameter of \( d_o = 1.2 \times 10^{-2} \, \text{m} \) and an inner diameter of \( d_i = 9 \times 10^{-3} \, \text{m} \). The needle wraps around the hernia ring, pulling and tightening the suture thread. Fig.~\ref{fig:hernia} shows the simulation at various stages. As the needle penetrates tissue and wraps around the hernia, the thread nodes follow, tightening around the ring.

In this simulation, the needle's position (represented by the lead node) is controlled in real-time by the user via keyboard arrows. The simulation is computed at the speed of 33Hz on a Dell Inspiron laptop, with specifications including 8GB RAM, and Intel i7 processor. Despite the modest capabilities of this system, the simulation performs efficiently, and the results remain realistic. With more advanced hardware typically used for real-time applications, the computational speed would be even higher, but this model demonstrates its realism and functionality even on basic consumer hardware.

\begin{algorithm}[t]
\caption{Real-Time Thread Sim with Visual Feedback}
\label{alg:haptic}
\begin{algorithmic}[1]
\State Initialize obstacles by smoothing edges and triangulating
\State Initialize states and reference velocity $x_i = x_i^0$, $v = v_i^0$
\State Get initial $h_{obs}$
\Loop
    \State Set \textit{$v_0$} $\gets$ User Input

    \State index $\gets i$ s.t. two elements of $h_{obs,i} <$ threshold
    \If{index is empty}
        \State Set \textit{Thread.Color} $\gets$ green 
    
    \ElsIf{$v_i \neq 0$}
            \State Decrease needle speed: $v_0 \gets v_0 \times\beta^{length(index)}$
            
           \State Set \textit{Thread.Color} $\gets$ Scale(red, length(index)) 
    \Else
            \State Set \textit{Thread.Color} $\gets$ blue 
    \EndIf
    
             \State Damp reference velocity $v_i \gets \kappa v_i$

    \State Determine constraints \eqref{eq:axb}
    \State Determine safe velocities $[u_i,u_0]$ according to \eqref{eq:control} %given reference velocities and current positions $v_i, v_0, x_i, x_0$

    \State Update dynamics according to \eqref{eq:dyn}
    %\State Update visuals with new positions and thread colors.
    
\EndLoop
\end{algorithmic}
\end{algorithm}

Visual feedback is implemented as described in Algorithm~\ref{alg:haptic}. As the needle moves, the obstacle avoidance barrier function for nodes approaching the tissue decreases towards zero. If the barrier function of node \(i\) approaches zero for two obstacles, the system detects a penetration path, triggering visual feedback to slow the needle, simulating increased resistance. Additionally, the needle turns orange to indicate friction, with the shade of orange and the degree of slowdown proportional to the number of nodes experiencing friction. If part of the thread is under friction but not being pulled, it turns blue. When the thread moves freely, it turns green. This visual feedback is entirely based on the thread's interaction with the environment, as captured by the barrier function values, without explicitly calculating forces. The barrier functions for this simulation are plotted over time in Fig.~\ref{fig:hobs}. 

Each node \(i = 1, \dots, n\) has velocity \(u_i\) and position \(x_i\). Needle velocity and position are represented by \(u_0\) and \(x_0\) respectively. Elements of $h_{obs,i}$ corresponds to the barrier function value of node \(i\) in respect to each obstacle. $0<\beta<1$ is a scaling factor for reducing the velocity based on friction. In the simulation here $\beta=0.9$ is used. When applying Equation~\eqref{eq:control}, the slack function weights are empirically  set as \( W_{\text{con}} = 10^5 \) and \( W_{\text{stiff}} = 1 \).

% \begin{algorithm}
% \caption{Constraint Calculation}
% \label{alg:cbf}
% \begin{algorithmic}[1]
% \Require Reference thread velocity $v_i$, Thread Position $x_i$,  Needle velocity $u_0$, Needle position $x_0$

% \Ensure Safe velocity controller $u_i$, Proximity to 
% \State Initialize states and reference velocity $x_i = x_i^0$, $k = u_i^0$

% \Loop
%     \State \textit{$u_0$} $\gets$ User Input
    
%     \State $[u_i, u_0, h_{obs}] \gets \texttt{CBF(}u_i, u_0, x_i, x_0\texttt{)}$

%     index $\gets i$ s.t. two elements of $h_{obs,i} <$ threshold
%     \If{index is empty}
%         \State Set \textit{Thread.Color} $\gets$ green 
    
%     \ElsIf{$u_i >$ velthresh}
%             \State $u_0 = u_0 \times\beta^{length(index)}$
%             \State $u_i = u_i \times \beta^{length(index)}$
%             \State Set \textit{Thread.Color} $\gets$ Scale(red, length(index)) 
%     \Else
%             \State Set \textit{Thread.Color} $\gets$ blue 
%     \EndIf
    
% \EndLoop
% \end{algorithmic}
% \end{algorithm}

\section{CONCLUSIONS}

In this work, we presented a novel approach to modeling cable-like objects, specifically suture threads, that is both computationally efficient and accurate. By utilizing control barrier functions and optimization techniques, our method avoids solving complex equations of motion, enabling real-time applications in surgical robotics and training simulations. Validation against experimental data, particularly with silk sutures, highlights its potential to advance autonomous surgical tasks. Future work will focus on extending the model to handle more complex procedures, including thread self-collision, such as suturing and ligation tasks.

\addtolength{\textheight}{-0cm}   % This command serves to balance the column lengths

%%%%%%%%%%%%%%%%%%%%%%%%%%%%%%%%%%%%%%%%%%%%%%%%%%%%%%%%%%%%%%%%%%%%%%%%%%%%%%%%

%%%%%%%%%%%%%%%%%%%%%%%%%%%%%%%%%%%%%%%%%%%%%%%%%%%%%%%%%%%%%%%%%%%%%%%%%%%%%%%%

%%%%%%%%%%%%%%%%%%%%%%%%%%%%%%%%%%%%%%%%%%%%%%%%%%%%%%%%%%%%%%%%%%%%%%%%%%%%%%%%

\section*{ACKNOWLEDGMENT}

The authors would like to thank Daniel (Qinhan) Wang for providing videos of the MagnetoSuture system.

%%%%%%%%%%%%%%%%%%%%%%%%%%%%%%%%%%%%%%%%%%%%%%%%%%%%%%%%%%%%%%%%%%%%%%%%%%%%%%%%

\bibliographystyle{IEEEtran}
%\bibliography{references}

\begin{thebibliography}{10}
\providecommand{\url}[1]{#1}
\csname url@rmstyle\endcsname
\providecommand{\newblock}{\relax}
\providecommand{\bibinfo}[2]{#2}
\providecommand\BIBentrySTDinterwordspacing{\spaceskip=0pt\relax}
\providecommand\BIBentryALTinterwordstretchfactor{4}
\providecommand\BIBentryALTinterwordspacing{\spaceskip=\fontdimen2\font plus
\BIBentryALTinterwordstretchfactor\fontdimen3\font minus \fontdimen4\font\relax}
\providecommand\BIBforeignlanguage[2]{{%
\expandafter\ifx\csname l@#1\endcsname\relax
\typeout{** WARNING: IEEEtran.bst: No hyphenation pattern has been}%
\typeout{** loaded for the language `#1'. Using the pattern for}%
\typeout{** the default language instead.}%
\else
\language=\csname l@#1\endcsname
\fi
#2}}

\bibitem{reddy2023advancements}
K.~Reddy, P.~Gharde, H.~Tayade, M.~Patil, L.~S. Reddy, and D.~Surya, ``Advancements in robotic surgery: a comprehensive overview of current utilizations and upcoming frontiers,'' \emph{Cureus}, vol.~15, no.~12, 2023.

\bibitem{attanasio2021autonomy}
A.~Attanasio, B.~Scaglioni, E.~De~Momi, P.~Fiorini, and P.~Valdastri, ``Autonomy in surgical robotics,'' \emph{Annual Review of Control, Robotics, and Autonomous Systems}, vol.~4, no.~1, pp. 651--679, 2021.

\bibitem{ostrander2024current}
B.~T. Ostrander, D.~Massillon, L.~Meller, Z.-Y. Chiu, M.~Yip, and R.~K. Orosco, ``The current state of autonomous suturing: a systematic review,'' \emph{Surgical Endoscopy}, vol.~38, no.~5, pp. 2383--2397, 2024.

\bibitem{lv2020review}
N.~Lv, J.~Liu, H.~Xia, J.~Ma, and X.~Yang, ``A review of techniques for modeling flexible cables,'' \emph{Computer-Aided Design}, vol. 122, p. 102826, 2020.

\bibitem{jourdes2022visual}
F.~Jourdes, B.~Valentin, J.~Allard, C.~Duriez, and B.~Seeliger, ``Visual haptic feedback for training of robotic suturing,'' \emph{Frontiers in Robotics and AI}, vol.~9, p. 800232, 2022.

\bibitem{yu2020real}
P.~Yu, J.~Pan, H.~Qin, A.~Hao, and H.~Wang, ``Real-time suturing simulation for virtual reality medical training,'' \emph{Computer Animation and Virtual Worlds}, vol.~31, no. 4-5, p. e1940, 2020.

\bibitem{moore2023interactive}
J.~Moore, H.~Scheirich, S.~Jadhav, A.~Enquobahrie, B.~Paniagua, A.~Wilson, A.~Bray, G.~Sankaranarayanan, and R.~B. Clipp, ``The interactive medical simulation toolkit (imstk): an open source platform for surgical simulation,'' \emph{Frontiers in Virtual Reality}, vol.~4, p. 1130156, 2023.

\bibitem{qi2017virtual}
D.~Qi, K.~Panneerselvam, W.~Ahn, V.~Arikatla, A.~Enquobahrie, and S.~De, ``Virtual interactive suturing for the fundamentals of laparoscopic surgery (fls),'' \emph{Journal of biomedical informatics}, vol.~75, pp. 48--62, 2017.

\bibitem{xu2018real}
L.~Xu and Q.~Liu, ``Real-time inextensible surgical thread simulation,'' \emph{International Journal of Computer Assisted Radiology and Surgery}, vol.~13, pp. 1019--1035, 2018.

\bibitem{ames2019control}
A.~D. Ames, S.~Coogan, M.~Egerstedt, G.~Notomista, K.~Sreenath, and P.~Tabuada, ``Control barrier functions: Theory and applications,'' in \emph{2019 18th European control conference (ECC)}.\hskip 1em plus 0.5em minus 0.4em\relax IEEE, 2019, pp. 3420--3431.

\bibitem{jankovic2021collision}
M.~Jankovic and M.~Santillo, ``Collision avoidance and liveness of multi-agent systems with cbf-based controllers,'' in \emph{2021 60th IEEE Conference on Decision and Control (CDC)}.\hskip 1em plus 0.5em minus 0.4em\relax IEEE, 2021, pp. 6822--6828.

\bibitem{liu2023clf}
J.~Liu, M.~Li, J.~W. Grizzle, and J.-K. Huang, ``Clf-cbf constraints for real-time avoidance of multiple obstacles in bipedal locomotion and navigation,'' in \emph{2023 IEEE/RSJ International Conference on Intelligent Robots and Systems (IROS)}.\hskip 1em plus 0.5em minus 0.4em\relax IEEE, 2023, pp. 10\,497--10\,504.

\bibitem{chinelato2023design}
C.~I.~G. Chinelato, B.~A. Ang{\'e}lico, J.~F. Justo, and A.~A.~M. Lagan{\'a}, ``Design of adaptive cruise control with control barrier function and model-free control,'' \emph{Journal of Control, Automation and Electrical Systems}, vol.~34, no.~3, pp. 470--483, 2023.

\bibitem{bruggemann2022simultaneous}
S.~Br{\"u}ggemann, D.~Steeves, and M.~Krstic, ``Simultaneous lane-keeping and obstacle avoidance by combining model predictive control and control barrier functions,'' in \emph{2022 IEEE 61st Conference on Decision and Control (CDC)}.\hskip 1em plus 0.5em minus 0.4em\relax IEEE, 2022, pp. 5285--5290.

\bibitem{sastry2013nonlinear}
S.~Sastry, \emph{Nonlinear systems: analysis, stability, and control}.\hskip 1em plus 0.5em minus 0.4em\relax Springer Science \& Business Media, 2013, vol.~10.

\bibitem{mair2020magnetosuture}
L.~O. Mair, X.~Liu, B.~Dandamudi, K.~Jain, S.~Chowdhury, J.~Weed, Y.~Diaz-Mercado, I.~N. Weinberg, and A.~Krieger, ``Magnetosuture: Tetherless manipulation of suture needles,'' \emph{IEEE transactions on medical robotics and bionics}, vol.~2, no.~2, pp. 206--215, 2020.

\bibitem{li2018formally}
A.~Li, L.~Wang, P.~Pierpaoli, and M.~Egerstedt, ``Formally correct composition of coordinated behaviors using control barrier certificates,'' in \emph{2018 IEEE/RSJ International Conference on Intelligent Robots and Systems (IROS)}.\hskip 1em plus 0.5em minus 0.4em\relax IEEE, 2018, pp. 3723--3729.

\bibitem{jamakovic2008robustness}
A.~Jamakovic and P.~Van~Mieghem, ``On the robustness of complex networks by using the algebraic connectivity,'' in \emph{International conference on research in networking}.\hskip 1em plus 0.5em minus 0.4em\relax Springer, 2008, pp. 183--194.

\bibitem{ferraguti2020control}
F.~Ferraguti, M.~Bertuletti, C.~T. Landi, M.~Bonf{\`e}, C.~Fantuzzi, and C.~Secchi, ``A control barrier function approach for maximizing performance while fulfilling to iso/ts 15066 regulations,'' \emph{IEEE Robotics and Automation Letters}, vol.~5, no.~4, pp. 5921--5928, 2020.

\bibitem{zeng2021safety}
J.~Zeng, B.~Zhang, and K.~Sreenath, ``Safety-critical model predictive control with discrete-time control barrier function,'' in \emph{2021 American Control Conference (ACC)}.\hskip 1em plus 0.5em minus 0.4em\relax IEEE, 2021, pp. 3882--3889.

\bibitem{fortune2017voronoi}
S.~Fortune, ``Voronoi diagrams and delaunay triangulations,'' in \emph{Handbook of discrete and computational geometry}.\hskip 1em plus 0.5em minus 0.4em\relax Chapman and Hall/CRC, 2017, pp. 705--721.

\end{thebibliography}

  % Replace 'main' with your actual .bbl filename

\end{document}